\documentclass[letterpaper, 10 pt, journal, twoside]{IEEEtran}
\usepackage{graphicx}
\usepackage{amsmath}
\usepackage{amssymb}
\usepackage{makecell}
\usepackage{multirow}
\usepackage{multicol}
\usepackage{caption}
\usepackage{algorithm}
\usepackage{algorithmic}
\usepackage{hyperref}

\ifCLASSINFOpdf
\else
\fi
\begin{document}
\title{GoalGrasp: Grasping Goals in Partially Occluded Scenarios without Grasp Training}

% \author{Michael~Shell,~\IEEEmembership{Member,~IEEE,}
%         John~Doe,~\IEEEmembership{Fellow,~OSA,}
%         and~Jane~Doe,~\IEEEmembership{Life~Fellow,~IEEE}% <-this % stops a space
% \thanks{M. Shell was with the Department
% of Electrical and Computer Engineering, Georgia Institute of Technology, Atlanta,
% GA, 30332 USA e-mail: (see http://www.michaelshell.org/contact.html).}% <-this % stops a space
% \thanks{J. Doe and J. Doe are with Anonymous University.}% <-this % stops a space
% \thanks{Manuscript received April 19, 2005; revised August 26, 2015.}}
\author{ Shun Gui*, Kai Gui* and Yan Luximon
\thanks{
The work was partially supported by a grant from the Research Grants Council of the Hong Kong Special Administrative Region, China (Project No. GRF/PolyU 15607922) and from Non-PAIR Research Centres of The Hong Kong Polytechnic University (Project No. P0052989). (Corresponding author: Yan Luximon)}
\thanks{Shun Gui and Yan Luximon are with the School of Design, The Hong Kong Polytechnical University. {\tt\small shun.gui@connect.polyu.hk;
yan.luximon@polyu.edu.hk}}
\thanks{Kai Gui is with the National Graduate College for Elite Engineers, Southeast University, Nanjing, China. {\tt\small by1707113@buaa.edu.cn; 103200203@seu.edu.cn}}
\thanks{* indicates the authors contribute equally.}}
\maketitle
%\thispagestyle{empty}
%\pagestyle{empty}

\begin{abstract}
Grasping user-specified objects is crucial for robotic assistants; however, most current 6-DoF grasp detection methods are object-agnostic, making it challenging to grasp specific targets from a scene. To achieve that, we present GoalGrasp, a simple yet effective 6-DoF robot grasp pose detection method that does not rely on grasp pose annotations and grasp training. By combining 3D bounding boxes and simple human grasp priors, our method introduces a novel paradigm for robot grasp pose detection. GoalGrasp's novelty is its swift grasping of user-specified objects and partial mitigation of occlusion issues. The experimental evaluation involves 18 common objects categorized into 7 classes. Our method generates dense grasp poses for 1000 scenes. We compare our method's grasp poses to existing approaches using a novel stability metric, demonstrating significantly higher grasp pose stability. In user-specified robot grasping tests, our method achieves a 94\% success rate, and 92\% under partial occlusion.
\end{abstract}

% \begin{IEEEkeywords}
% IEEE, IEEEtran, journal, \LaTeX, paper, template.
% \end{IEEEkeywords}

% \fi
\IEEEpeerreviewmaketitle

\section{INTRODUCTION}

\section{Introduction}
\IEEEPARstart{I}{n} human-robot interaction scenarios, robots often rely on receiving instructions from humans to perform specific manipulations. For instance, in the case of home service robots, it is common for users to provide the robot with a specific goal, such as grasping a particular object [1]. The ability to grasp user-specified objects is significant in human-robot interaction. This capability is particularly essential for individuals with limited mobility, such as elderly individuals or patients, where robots can provide fundamental assistance [2].

However, the state-of-the-art research [3] in robotic grasping often focuses less on user-specified grasping. In many studies, the object for the robot to grasp is unknown, meaning that the robot is unaware of the object it grasps [4,5]. These studies typically involve detecting grasp poses for the entire scene and then selecting the best pose for execution. While this mode is suitable for scenarios like bin picking [6], it may not adequately meet the requirements of user-driven human-robot interactions in daily life. We consider that extending such methods to target-oriented grasping tasks presents significant challenges, including (1) the inability to categorize generated grasp poses according to the object's class and (2) the diverse nature of scenarios encountered by robots, which often lack sufficient training data to ensure reliable grasp performance. One potential solution is to manually annotate a large-scale dataset of categorized grasp poses. However, this approach significantly limits the efficiency of grasp robotics applications due to the labor-intensive and time-consuming process of manually labeling 6D annotations [7,8]. Considering the wide range of application scenarios for robots, this method proves to be inefficient.

Some target-oriented grasping research [9,10] begins with object detection or instance segmentation [11] in images and then detects the corresponding grasp poses in the associated point cloud to achieve the user-specified grasping. However, these methods are susceptible to interference from nearby objects as it may inadvertently included neighboring objects during target object detection. Additionally, the performance diminishes greatly when the target object is partially occluded. Indeed, occlusion poses a significant challenge to the performance of existing robotic grasping methods [12]. However, in target-driven grasping scenarios, where objects are partially occluded, it is unavoidable. Therefore, overcoming the issue of occlusion is one of the difficulties that our research needs to address.

In this work, we propose a novel 'object-level' grasp pose generation method called GoalGrasp, which enables the grasping of user-specified objects even in the presence of partial occlusion. Unlike existing learning-based approaches, our method eliminates the need for grasp-specific training, thereby alleviating the requirement for manual annotation of 6D grasp poses. This significantly enhances the efficiency of our method when applied to new scenarios. An essential characteristic of our method is its 'object-level' granularity, ensuring the generation of grasp poses even in the presence of partial occlusion. To achieve 'object-level' grasp pose detection, we first employ a 3D object detection method called Recursive Cross-View (RCV) [13] to obtain 3D bounding boxes for objects within the scene. 

Through reflection on human grasping behavior, we observe that human grasping abilities can also be considered 'object-level'. For a given object, humans often rely on simple heuristics to achieve successful grasps. For example, when grasping a box, our heuristic is to grasp two opposing faces, while for an apple, our heuristic is to grasp two points along its 'diameter'. We do not consider grasping a single vertex of a box or two adjacent faces as effective grasping strategies. We find that these simple grasping heuristics can greatly simplify grasp pose generation methods, reducing the need for complex learning-based models and enabling rapid robotic object grasping. Utilizing the 3D bounding box, object category, and some simple grasping heuristics, we propose a straightforward yet effective 'object-level' grasp pose generation method.

We conduct four experiments. Firstly, we category the 18 objects within the collected dataset into seven major classes. Corresponding grasp pose generation algorithms are designed for each class, enabling the generation of dense grasp poses for the objects in these 1000 scenarios. To evaluate the quality of the grasp poses, we introduce a novel stability metric and use it to compare the grasp poses generated by our method against those generated by state-of-the-art approaches. The experimental results demonstrate that our method significantly outperforms existing methods. To ensure a fair comparison, the existing methods are not retrained. As our method is 'object-level', the generated grasp poses are associated with specific object categories, enabling direct execution of user-specified object grasping. We deploy GoalGrasp on a real robot and perform 500 user-specified target grasping tasks, achieving a grasp success rate of 94\%. Moreover, we conduct 100 user-specified target grasping tasks in scenes with occlusions, resulting in a grasp success rate of 92\%. 

Our main contributions are summarized as follows:
\begin{itemize}
    \item We propose an 'object-level' 6-DoF grasp pose generation method that eliminates the need for any grasp-specific training. This method enables grasping of user-specified targets and can adapt to scenes with occlusions.
    \item We propose a novel stability metric to evaluate the grasp poses associated with the object. Using this metric, we compare our method with existing approaches, which demonstrates that our method outperforms them significantly.
    \item We conduct extensive experiments on a real robot, which involve grasping user-specified objects as well as partially occluded objects.
\end{itemize}

\section{Related Work}
\subsection{Learning-based Grasp Pose Detection}
Recent advancements in deep learning have paved the way for the development of data-driven systems in robotic grasping [14]. However, learning-based robotic grasping methods often require large-scale annotated datasets for training [3].
However, annotating these datasets can be time-consuming and labor-intensive [7,8], which to some extent limits the rapid application of learning-based methods in various real-world robotic scenarios. [3] showed the drawback of the frequently adopted sim2real methods in the grasping community.
Furthermore, learning-based methods for robotic grasping often involve the design and training of sophisticated neural network models to predict grasp poses [3,15,16,17]. Mousavian et al. [18] leveraged a variational auto-encoder to generate a set of grasps, which were then evaluated and refined using a grasp evaluator model. 
Wang et al. [19] developed an end-to-end network named GSNet, which combines a graspness model to predict grasp poses. Fang et al. [20] proposed a grasp pose prediction network that learns approaching direction and operation parameters separately, using point cloud inputs. One of the challenges with learning-based methods is that they can be difficult to generalize to new scenarios and objects. In many cases, we need to recollect data, label it, and then retrain the model, which can be time-consuming and inefficient, especially for diverse robotic applications. Our method leverages some simple grasping priors to significantly enhance the efficiency.

\subsection{Target-oriented Grasp Pose Detection}
Many existing grasp pose detection methods operate by taking a scene as input and generating multiple grasp poses. Then the robot selects the optimal grasp poses for execution [3,17]. While this paradigm is suitable for scenarios such as bin picking, it may not directly address the requirement of grasping user-specified objects in human-robot interaction settings. Recognizing this limitation, few research efforts have specifically focused on the task of grasping user-specified objects [9,10,21,22]. Murali et al. [9] utilized instance segmentation [11] to align grasps with target objects. Liu et al. [10] leveraged semantic segmentation module to locate the target first, and then predicted the grasp poses. The majority of these studies concentrated on 3-DoF grasping, which produces grasping poses within the camera plane, restricting their practicality. Furthermore, they have not evaluated their performance in situations where the target object is partially obscured.

\subsection{Grasp Pose Detection in Partially Occluded Scenarios}
One category of grasp pose detection methods is 6D pose estimation-based [8,23,24,25,26]. Deng et al. [8] proposed a self-supervised 6D object pose estimation for robotic grasping. Zhang et al. [25] presented a practical grasping approach for robots that utilizes 6D pose estimation along with corrective adjustments for protection. However, these methods have not demonstrated satisfactory performance in scenarios where the target objects are partially occluded, since 6D pose estimation methods is susceptible to occlusions [27]. The same issue also arises in methods for detecting graspable rectangles [17,28]. 
Some methods [18,29,30] adopted a sampling-evaluation strategy, which involves sampling potential grasp candidates on point cloud data and then assessing their quality using a neural network. In addition, some studies [3,19] employed neural networks to regress grasp poses on point clouds. However, for these point cloud-based methods, they struggle to generate accurate grasp poses when the objects are partially occluded, especially when the optimal grasp regions are occluded. The nature of these methods limits their ability to generate grasp poses in regions where point cloud data is unavailable. Additionally, they often lack a holistic understanding of the grasped objects.

\section{Problem Definition}
Our primary objective is to enable robots with a two finger parallel-jaw gripper, to grasp target objects from a scene according to instructions from a human.
Specifically, we represent the grasp formulation for the robot as

\begin{equation}
\small
    \textit{G} = [~\textbf{R} ~~~\textbf{T} ~~~\textit{c}~~~\textit{s}~~~\textbf{C}~]
\end{equation}
s.t.
\[ \begin{cases}
\small
    s = E(\mathbf{R}, ~\mathbf{T}, ~B) \\
    \mathbf{C} \leftarrow \{w: \text{width};~ d: \text{depth}\}
\end{cases}
\]
where $\mathbf{R}\in \mathbb{R}^{3\times3}$ demonstrates the orientation of the gripper, $\mathbf{T}\in \mathbb{R}^{3\times1}$ denotes the location of the gripper, $c$ refers to the gripper category, $s$ refers to the score of the grasp pose evaluated by $E(\mathbf{R}, ~\mathbf{T}, ~B)$, $B$ is the 3D box of the target object, and \textbf{C} is the configuration, including the width ($w$) and depth ($d$) for the grasp pose. Therefore, we generate a set of grasp poses, $\textbf{G}(c) = \{G_1, G_2, ..., G_n\}$, for one object, and generate several sets of grasp poses for a scene. 

\section{3D Detection on New Objects}

The method we propose is based on utilizing 3D bounding boxes to detect the grasp poses of target objects, which requires to efficiently perform 3D object detection for various items in new scenes. However, the majority of existing 3D object detection methods heavily rely on extensive manual annotation of 3D labels. Clearly, manual annotation of 3D labels is both labor-intensive and time-consuming, making it impractical for achieving efficient robotic grasping. In contrast, Recursive Cross-View (RCV) introduced a completely 3D label-free 3D object detection method that solely utilizes easily obtainable 2D bounding box labels to achieve 3D detection of novel objects. Next, we describe how to employ this method to accomplish 3D detection of tabletop objects.

To achieve 3D tabletop object detection in new scenes without relying on 3D annotations, we employ an Azure Kinect DK to gather RGB images and point cloud data. Initially, we manually annotate 2D bounding boxes for RCV. Subsequently, we develop a 3D detector that operates without direct access to 3D annotations throughout the process. Figure 1 illustrates the process of 2D labeling and inferring 3D bounding boxes from the collected data. First, we manually label 2D bounding boxes on the raw RGB images captured by the Kinect (Figure 1, first row). These labeled bounding boxes are then utilized to train a 2D detector, enabling 2D object detection on the RGB images. Next, we filter out points located outside the 2D bounding boxes and project the remaining points orthogonally, resulting in two images (Figure 1, second row). Manual labeling is performed on these two images, generating two red 2D bounding boxes. Similarly, we filter out the points outside these newly labeled bounding boxes and project the remaining points orthogonally, producing two new images (Figure 1, fourth row). We annotate these two images with two red 2D bounding boxes. Importantly, all 3D bounding boxes are inferred based on the previously labeled 2D bounding boxes rather than relying on manual annotations. Subsequently, we utilize the 2D bounding boxes to train a 2D object detector (YOLOv5), which is then employed to develop an RCV model capable of detecting tabletop objects. For further details on the RCV model, please refer to [13].

In this study, we specifically select 18 different objects and annotated about 200 2D bounding boxes for each object. The annotation process for each object took approximately half an hour with one annotator. Leveraging these 2D annotations, we realize 3D detection. The RCV method, relying solely on established 2D detection techniques, demonstrates excellent robustness. We utilize the RCV to construct a dataset comprising 1,000 scenes, encompassing 15 categories and 18 different objects, with over 5,000 3D bounding boxes. Some samples are demonstrated in Figure 2. Subsequently, we employ a grasp generation strategy to generate dense grasp poses for each object in the scenes.

\begin{figure}[!t]
\centering
\includegraphics[width=0.65\linewidth]{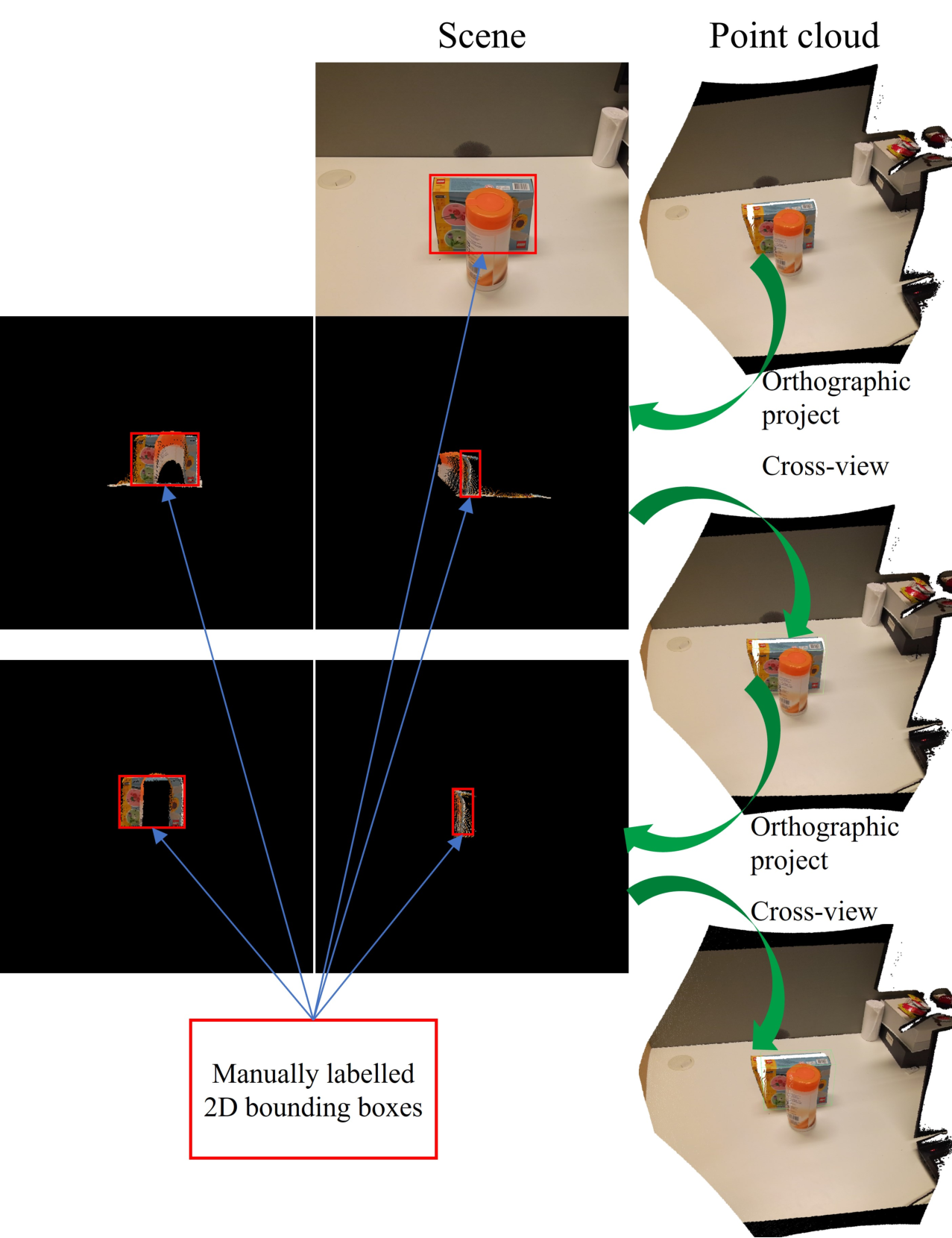}
\caption{2D labeling and inferring 3D bounding boxes for RCV on the collected data.}
\label{fig_2}
\end{figure}

\begin{figure}[!t]
\centering
\includegraphics[width=0.7\linewidth]{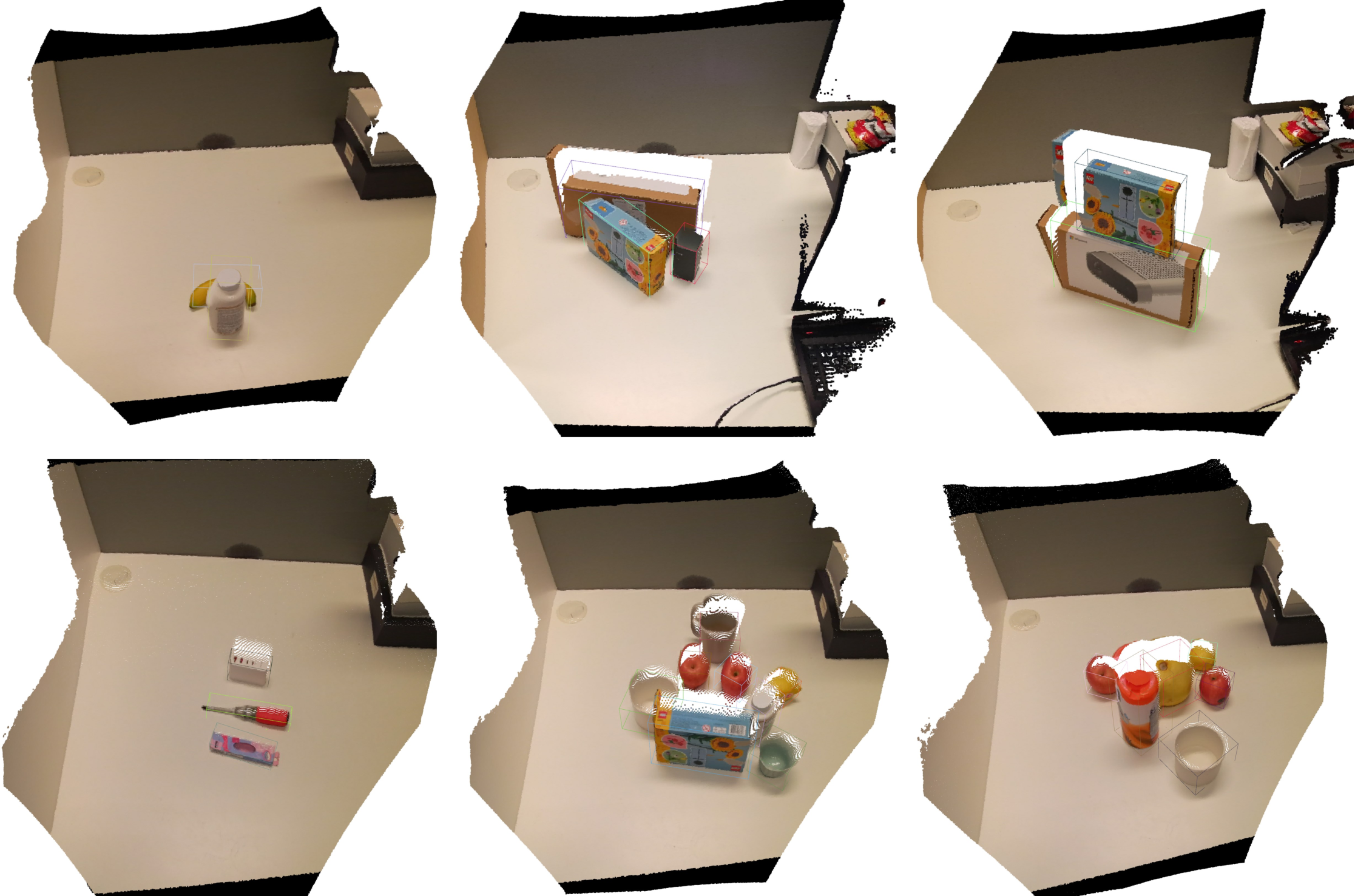}
\caption{3D-TABLETOP-OBJECT dataset with 15 categories: large box, small box, large cylinder, small cylinder, bowl, cucumber, banana, tape, screw, apple, lemon, grapefruit, pen, jar, mug.}
\label{fig_2}
\end{figure}

\section{Grasp Strategy}

In this section, we propose grasping pose generation strategies for the objects in the 1000 scenes mentioned in Section IV. Note that we use these objects as examples, many items not included in the dataset can also leverage these strategies. Here, we classify these items based on their geometric characteristics into box-shaped, spherical, cylindrical, curved, container, tool, and ring objects. Firstly, we summarize human's grasping priors specific to each object category. Subsequently, leveraging this prior knowledge and the 3D bounding boxes in the object coordinate system, we design a grasp generation strategy to obtain a dense set of grasp poses.

\subsection{Box-shaped Objects}
In everyday life, there are various box-shaped objects, such as packaging boxes and power banks. Regarding these objects with box-like shapes, human grasping prior knowledge can be summarized in the absence of considering the maximum gripper width. Specifically, the following priors apply: (1) The grasp position is located along the symmetric axis of each face, (2) the gripper's depth direction aligns closely with the normal orientation of each face, (3) the gripper's width direction aligns with the corresponding edge direction, and (4) the grasp width and depth correspond to the respective edge lengths. For (1), the sampling trajectories (ST) of grasp positions are shown as the red lines in Figure 3(a). Mathematically, we express them as 

$\small ST_0$ $\leftarrow$
$
\small
\left\{
\begin{aligned}
  %&GS_1 = 
  \begin{cases}
    x = x_{mid} \\
    y = 0 \\
    z = t,~ t\in z_{range}
  \end{cases}
  %\quad,
  ,
  %&GS_2 = 
  \begin{cases}
    x = x_{mid} \\
    y = y_{max} \\
    z = t,~ t\in z_{range}
  \end{cases}
  ,
\end{aligned}
\right.
$

$
\small
~~~~~~~~~~\left.
\begin{aligned}
  %&GS_1 = 
  \begin{cases}
    x = 0 \\
    y = y_{mid} \\
    z = t,~ t\in z_{range}
  \end{cases}
  %\quad,
  ,
  %&GS_2 = 
  \begin{cases}
    x = x_{max} \\
    y = y_{mid} \\
    z = t,~ t\in z_{range}
  \end{cases}
  ,
\end{aligned}
\right.
$

$
\small
~~~~~~~~~~\left.
\begin{aligned}
  %&GS_1 = 
  \begin{cases}
    x = x_{mid} \\
    y = t, ~ t\in y_{range}\\
    z = 0
  \end{cases}
  %\quad,
  ,
  %&GS_2 = 
  \begin{cases}
    x = x_{mid} \\
    y = t, ~ t\in y_{range} \\
    z = z_{max}
  \end{cases}
  ,
\end{aligned}
\right.
$

$
\small
~~~~~~~~~~\left.
\begin{aligned}
  %&GS_1 = 
  \begin{cases}
    x = 0 \\
    y = t, ~ t\in y_{range}\\
    z = z_{mid}
  \end{cases}
  %\quad,
  ,
  %&GS_2 = 
  \begin{cases}
    x = x_{max} \\
    y = t, ~ t\in y_{range} \\
    z = z_{mid}
  \end{cases}
  ,
\end{aligned}
\right.
$

$
\small
~~~~~~~~~~\left.
\begin{aligned}
  %&GS_1 = 
  \begin{cases}
    x = t, ~ t\in x_{range} \\
    y = y_{mid}\\
    z = 0
  \end{cases}
  %\quad,
  ,
  %&GS_2 = 
  \begin{cases}
    x = t, ~ t\in x_{range} \\
    y = y_{mid} \\
    z = z_{max}
  \end{cases}
  ,
\end{aligned}
\right.
$

$
\small
~~~~~~~~~~~\left.
\begin{aligned}
  %&GS_1 = 
  \begin{cases}
    x = t, ~ t\in x_{range} \\
    y = 0\\
    z = z_{mid}
  \end{cases}
  %\quad,
  ,
  %&GS_2 = 
  \begin{cases}
    x = t, ~ t\in x_{range} \\
    y = y_{max} \\
    z = z_{mid}
  \end{cases}
\end{aligned}
\right\}.
$

\textbf{Algorithm 1} presents the grasp pose generation algorithm for box-shaped objects. For each element in $ST_0$, we sample multiple grasp points and derive the corresponding grasp pose (line 8-19 and 30), width (line 21-26), and depth (line 28). Figure 3(b) illustrates the generated grasping poses. It is important to note that these grasping poses do not account for factors such as the maximum gripper width and environmental constraints, which will be discussed later.

\begin{algorithm}
    \small
    \caption{Generate grasp poses for box-shaped objects}
    \label{alg1}
    \begin{algorithmic}[1]
        \REQUIRE 3D bounding box $B$, point cloud of object $PC$, empty buffer $g$, empty buffer $W$, empty buffer $D$, sampling quantity $N$, maximum gripper depth $gd$.
        \REQUIRE $x\_l, y\_l, z\_l$ of $B$. $x_{min},$ $ x_{max},$ $ y_{min},$ $ y_{max}, $ $z_{min}, $ $z_{max}$ are the minimum and maximum values of B in the coordinate system of the object.
        \STATE $CT.x = (x_{min}+x_{max}) / 2$
        \STATE $CT.y = (y_{min}+y_{max}) / 2$
        \STATE $CT.z = (z_{min}+z_{max}) / 2$
        \FOR{$st$ in $ST_0$}
            \STATE // generate grasp poses
            %\STATE sample $N$ points $\{n_1, n_2, ..., n_N\}$ on $st$
            \FOR{$i\leftarrow 0$ to $N$}
            \STATE Randomly sample a point $p$ on $st$

            \STATE Calculate the $X$, $Y$, and $Z$ of the gripper: $X$ is from $p$ to the inside of $B$; $Y$ is perpendicular to $st$ and $X$; $Z$ is $X$ $\times$ $Y$.

            \STATE Grasp width $w=max(|x\_l*[1,0,0]\times Y|,\ |y\_l*[0,1,0]\times Y|,\ |z\_l*[0,0,1]\times Y|)$
            
            \STATE Grasp depth $d=min(x\_l/2,~y\_l/2, $
            \STATE$~z\_l/2, ~gd)$
            %\STATE $Z = X$ $\times$ $Y$
            \STATE $R = [X, Y, Z]$
            \STATE $T = p$
            \STATE
            \STATE $g\leftarrow$ $g$ $\bigcup$ $\begin{bmatrix}R & T \\ \mathbf{0} & 1 \\ \end{bmatrix}$ ~$W\leftarrow$ $W$ $\bigcup$ $w$ ~ $D\leftarrow$ $D$ $\bigcup$ $d$
            \STATE Rotate $R$ along $Y$ with a random angle in $[-1/4\pi, 1/4\pi]$
            \STATE
            \STATE $g\leftarrow$ $g$ $\bigcup$ $\begin{bmatrix}R & T \\ \mathbf{0} & 1 \\ \end{bmatrix}$ ~$W\leftarrow$ $W$ $\bigcup$ $w$ ~ $D\leftarrow$ $D$ $\bigcup$ $d$
            \ENDFOR
        \ENDFOR
    \STATE Transfer $g$ to the robot coordinate system.
    \RETURN $g,~W,~ D$
    \end{algorithmic}  
\end{algorithm}

\begin{figure}[!t]
\center\includegraphics[width=0.8\linewidth]{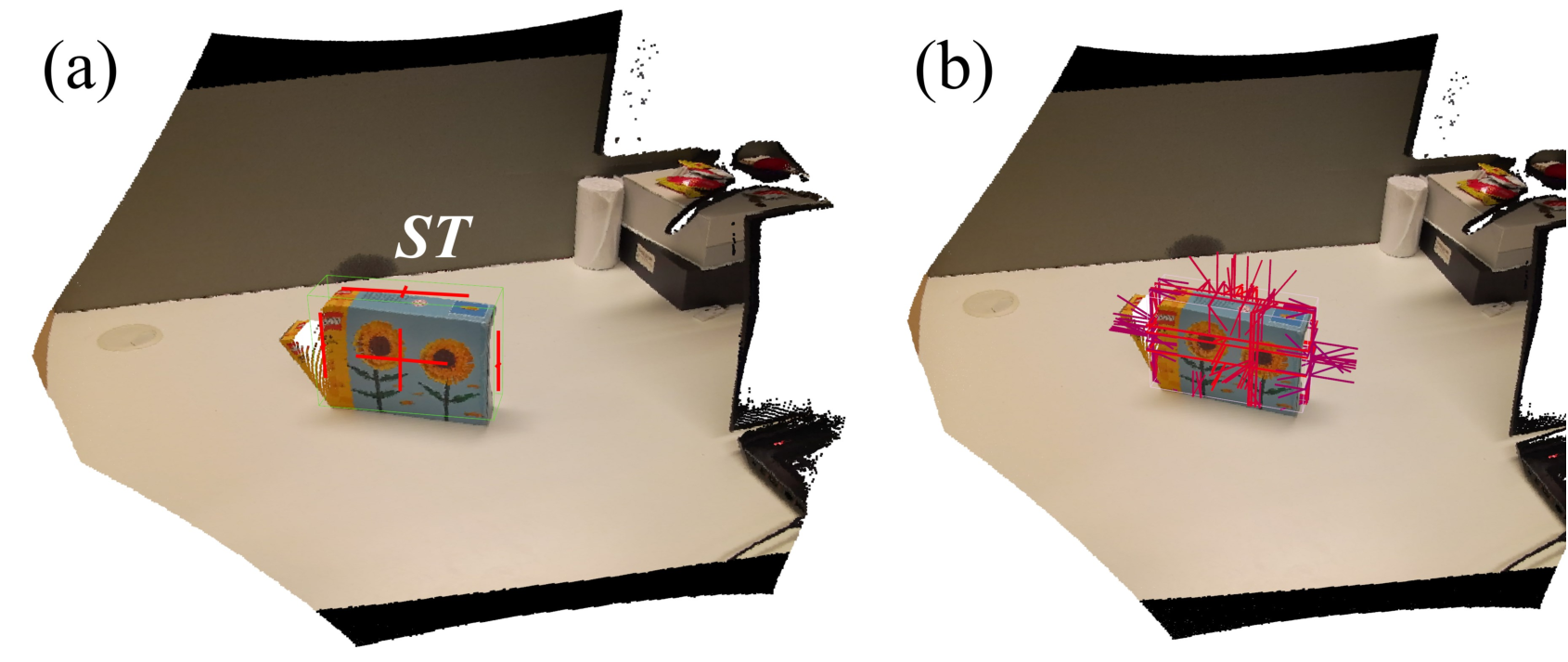}
\caption{Grasp poses generation for box-shaped objects. (a) Sampling trajectory (ST) for grasp points. (b) Generated grasp poses without filtering.}
\label{fig_2}
\end{figure}

\textbf{We can use similar steps for spherical, cylinder, curved, container, tool, and
Ring objects generate grasping poses. For the specific algorithm, see the Section II of the Supplementary Materials.}

\subsection{Filtering Metrics}
The grasp pose generation algorithm does not consider the environmental constraints and interferences between objects, resulting in the existence of infeasible grasp poses, such as those in close proximity to other objects or colliding with the tabletop. In this section, we propose the design of filtering metrics to remove these infeasible grasp poses. The first criterion is the orientation of the grasp pose, where valid grasp poses are either horizontal or inclined downward. This facilitates motion planning for the robot and prevents collisions with the tabletop. We measure this criterion using the cosine distance between the depth direction of the grasp pose and the direction of gravity, which is denoted as
\begin{equation}
\small
    dcos = 1-\frac{\boldsymbol{X} \cdot \boldsymbol{g}}{|\boldsymbol{X}||\boldsymbol{g}|}<TH_0
\end{equation}
where $\boldsymbol{X}$ is the depth direction of the grasp pose, $\boldsymbol{g}$ is the direction of gravity, which is $[0,1,0]$.T for our depth camera. $TH_0$ is the threshold.

The second criterion is that the grasp point must be positioned as a certain distance above the tabletop to avoid collisions between the robot and the tabletop. It can be represented as Eq.(3)
\begin{equation}
\small
    max(BBox[:,1])-G.T[1,0]>TH_1
\end{equation}
where $BBox\in \mathbb{R}^{8\times3}$ is the 3D bounding box of the object, $G.\mathbb{T} \in \mathbb{R}^{3\times1}$ is the location of the grasp pose. Note that the direction of gravity is defined as $[0, 1, 0]$.T. $TH_1$ is the threshold.

The third criterion is the maximum width of the gripper. It is evident that if the generated grasp pose's width $(w)$  exceeds the maximum gripper width $(w_{max})$, the grasp pose is considered invalid. This criterion can be represented as Eq.(4)
\begin{equation}
\small
    w_{max} - w>0
\end{equation}

The fourth criterion is to ensure that the grasp point is positioned at a distance greater than a certain threshold from other objects to avoid collisions between the robot and other objects. To determine the distance between the grasp pose and other objects, our method utilizes the 3D bounding boxes of other objects. We discretize a certain number of points on the surface of the bounding boxes and derive the minimum distance between the grasp pose and these points. If the minimum distance is greater than the predefined threshold, the grasp pose is considered valid. The criterion is shown as Eq.(5).
\begin{equation}
\small
\begin{split}
    min\{dist(G.T, ~BP_0), 
    &~dist(G.T, ~BP_1),\\
    ~dist(G.T, ~BP_2),~ ..., 
    &~dist(G.T, ~BP_n)\}>TH_2
 \end{split}   
\end{equation}
where $dist$ represents the Euclidean distance, $G.T\in \mathbb{R}^{3\times1}$ indicates the location of grasp pose, $BP\in \mathbb{R}^{1\times3}$ denotes a point on the 3D bounding boxes, and $n$ signifies the number of sampled points. $TH_2$ is the threshold.

We present four fundamental criteria here for filtering out invalid grasp poses. However, additional criteria can be extracted to ensure that the generated grasp poses meet the specific requirements of the given scenario.

\subsection{Evaluation Metric}
Leveraging the proposed grasp pose generation and filtering methods, we can obtain a dense set of feasible grasp poses. However, their performance lacks a quantifiable measure. Since our method does not rely on any grasp labels or training, some metrics previously used by other methods, such as AP, recall, and F1 score, are not applicable to our approach. To quantitatively evaluate the grasp poses generated by our method, we propose an object-level comprehensive metric to measure grasp stability. The fundamental principle of this metric is to encourage grasp points to be as close as possible to the object's center.
In [3], the awareness of the center of gravity (COG) for objects was introduced to assess the stability of grasp poses. Specifically, the normalized perpendicular distance (denoted as $d_1$) between gripper plane and the COG of the object is defined as the stable score. However, this metric  appears to be insufficient in certain cases, as demonstrated in the left subplot of Figure 4, where $d_1$ is zero, but the grasp appears to be unstable. We believe that this instability is due to the distance between the touch point and the COG, denoted as $d_2$. Therefore, we propose to incorporate both $d_1$ and $d_2$ to comprehensively evaluate the stability of grasps poses. The right subplot of Figure 4 illustrates a grasp pose's $d_1$ and $d_2$. Eq.(6) demonstrates the proposed evaluation metric.
\begin{equation}
\small
    M = \alpha \left[1-\frac{d_1}{l_{diag}/2} \right]+(1-\alpha) \left[1-\frac{d_2}{l_{diag}/2} \right]
\end{equation}
where $\alpha$ is a weight coefficient, $l_{diag}$ denotes the length of the diagonal of the 3D bounding box of the object. It is utilized to normalize $d_1$ and $d_2$ to the range $[0,~1]$. $d_1$ and $d_2$ are also derived using the 3D bounding box. Specifically, we consider the center of the 3D bounding box as the COG and then compute $d_1$ and $d_2$. Here, $\alpha$ is specified as 0.5.

\begin{figure}[!t]
\centering
\includegraphics[width=0.5\linewidth]{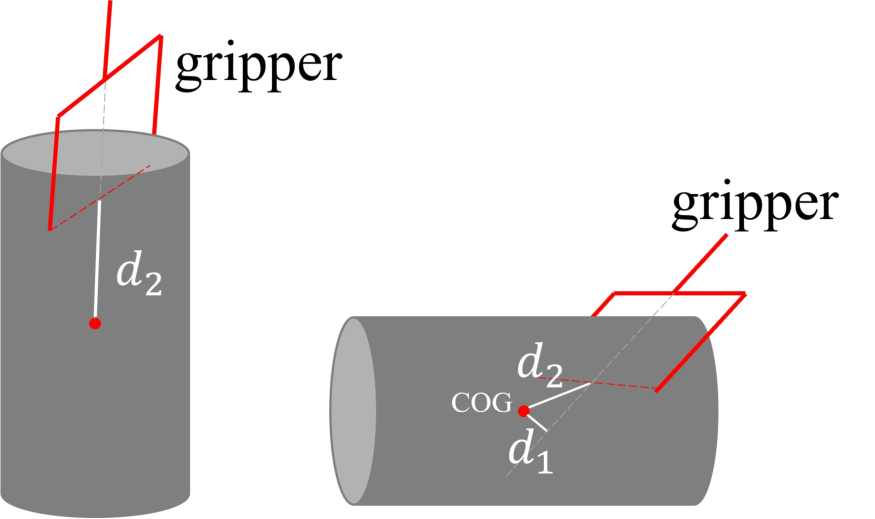}
\caption{The illustration of the novel stability metric.}
\label{fig_2}
\end{figure}

\begin{algorithm}
    \small
    \caption{Generate grasp poses for a scene}
    \label{alg1}
    \begin{algorithmic}[1]
        \REQUIRE Trained RCV; $img$, $PC\in \mathbb{R}^{n\times3}$ and $rgb \in \mathbb{R}^{n\times3}$ of the scene, empty buffer \textbf{G}.
        \STATE \textbf{boxes}, \textbf{labels}, \textbf{confs}, \textbf{obj\_pcs} = RCV$(img, PC, rgb)$
        \STATE // \textbf{obj\_pcs} is a set of segmented points for each objects.
        \FOR{$i \leftarrow$ 0 to \text{len(\textbf{boxes})-1}}
            \STATE Use \textbf{labels[i]} to retrieve the corresponding grasping pose generation algorithm, denoted as \textbf{GA}. 
            \STATE $g,~W,~ D$ = \textbf{GA}(\textbf{boxes[i]}, \textbf{obj\_pcs[i]})
            \STATE Filter $g,~W,~ D$ using Eq.(2)-(5).
            \STATE Utilize Eq.(6) to measure the grasp poses, and obtain the scores, denoted as $s$.
            \STATE $s$ = \textbf{confs[i]}$\times s$
            \STATE \textbf{G}$\leftarrow$ \textbf{G} $\bigcup$ $[g,~W,~ D, ~ s]$
            %\ENDFOR
        \ENDFOR
    \RETURN \textbf{G}
    \end{algorithmic}  
\end{algorithm}

\subsection{The Nature of Overcoming Occlusion}
Many existing robotic grasping studies are sampling-based, where grasp points are sampled from point clouds to generate grasp poses [3,19]. Additionally, to avoid collisions, grasp poses in contact with the point cloud are discarded. This approach can be considered 'point-level' grasp pose detection, but it fails to distinguish between points belonging to the object and noise, leading to the elimination of some grasp poses in contact with noise points. Moreover, this method performs poorly when objects are occluded and cannot sample grasp points in such cases. We believe that this 'point-level' approach lacks a holistic understanding of the grasping object. In contrast, our method is 'object-level', which addresses these problems to some extent. 

Intuitively, humans have the ability to achieve stable grasping even with only partial visibility of an object. This ability is based on human perceptual capabilities, such as the recognition of object categories and an understanding of their spatial extent. Our proposed method is directly inspired by this notion and leverages 3D object detection to enable robots to identify object categories and estimate their spatial occupancy, even in scenarios where objects are partially occluded. Figure 5 demonstrates that our method can detect the 3D bounding boxes of partially occluded objects. Subsequently, the 3D bounding box is combined with simple human grasping priors to generate 6D grasp poses. Compared to sampling-based methods, our approach can be regarded as a 3D bounding box-based method. This implies that our method does not need to consider the completeness of the point cloud of the object when generating grasp poses. Therefore, our method inherits the capability of overcoming partial occlusion from 3D bounding box detection.

The grasping capability of GoalGrasp is derived from RCV and human-inspired grasping heuristics. In our approach, RCV and human-inspired heuristics complement each other: RCV handles recognition, including overcoming occlusion issues, while human-inspired heuristics generate the grasp poses. To the best of our knowledge, our method is the first 6D grasp pose detection algorithm based on 3D bounding boxes. This also allows our method to better handle occlusions compared to existing methods, such as those based on 6D pose estimation or learning-based approaches.

\begin{figure}[!t]
\centering
\includegraphics[width=0.7\linewidth]{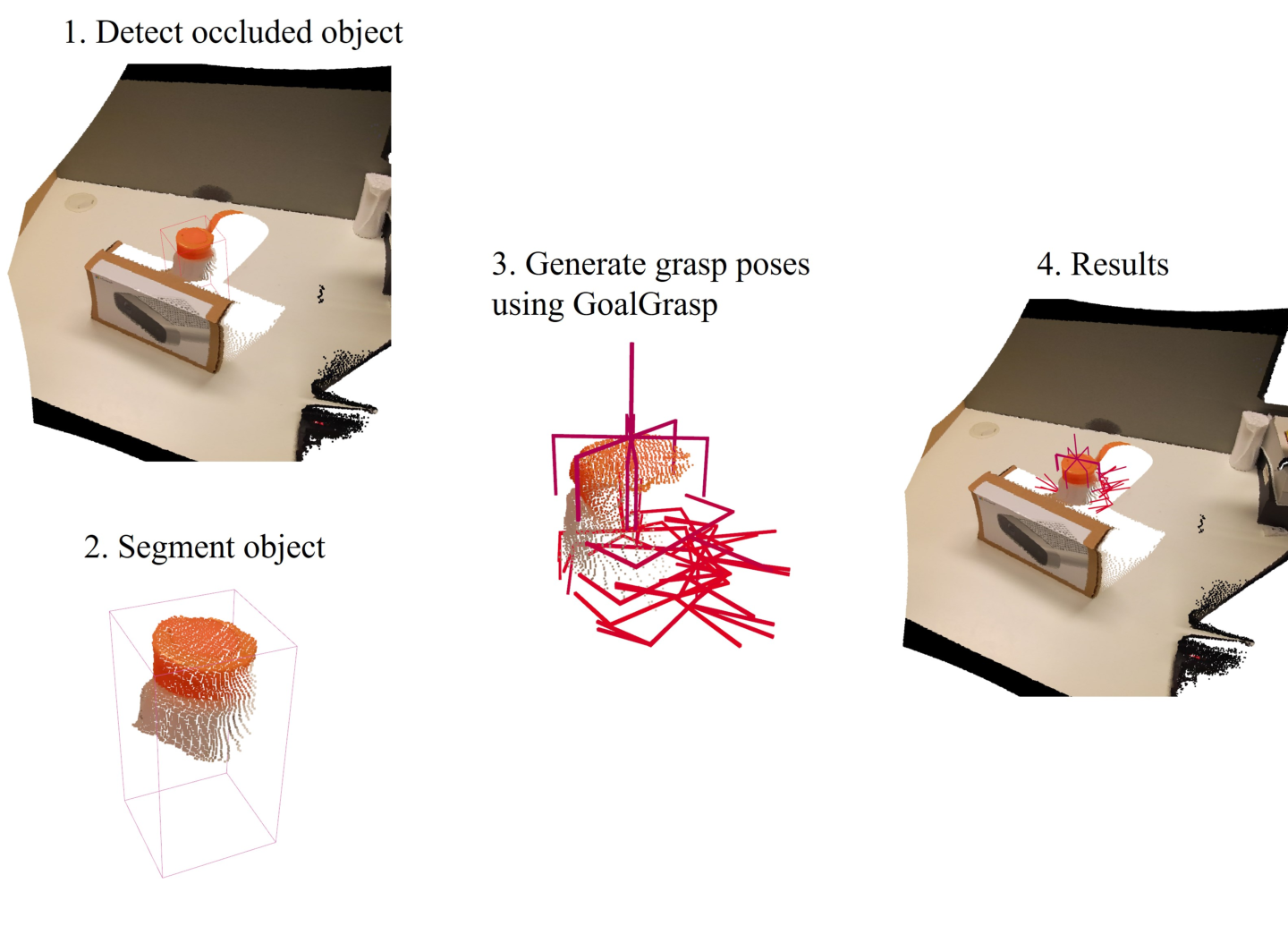}
\caption{The illustration of overcoming occlusion.}
\label{fig_2}
\end{figure}

\subsection{Generate Grasp Poses for A Scene}
In this section, we present a comprehensive algorithm for generating grasp poses in a given scene by integrating the previously mentioned grasp pose generation algorithms, filtering metrics, and evaluation metrics. The process begins by utilizing a pre-trained 3D detection model, RCV, to detect objects in the scene, which provides the 3D bounding boxes, labels, confidences, and point clouds for each detected object. Next, we retrieve the corresponding grasp pose generation algorithm based on the label to generate dense grasp poses for each object. These grasp poses then undergo filtering using Eq.(2)-(5) to eliminate invalid poses. Subsequently, the proposed evaluation metrics are applied to assess the grasp poses. To account for the impact of the 3D bounding box's quality on the grasp pose, we multiply the obtained scores by the confidence associated with each 3D bounding box. This adjusted score serves as the final evaluation score for each grasp pose. \textbf{Algorithm 2} illustrates the process of generating grasp poses for a given scene.

\section{Experiments}
\subsection{Generating Dense Grasp Poses for 1,000 Scenes}
Our proposed method enables object grasping without the need for any grasp training, which holds significant value in the diverse and dynamic scenarios of robotic grasping. Our method not only facilitates the rapid deployment of robotic grasping but also eliminates the need for laborious annotation of 6D grasp poses. To validate the effectiveness of our approach, we generate dense grasp poses for all objects in the 1000 scenes established in Section IV, utilizing \textbf{Algorithm 2}. In this dataset, there are 18 types of objects. We category these 18 types into 7 shape categories, each corresponding to a specific grasp pose generation algorithm, as shown in the Supplementary Materials (Table III). 
The last row of Figure 6 shows some generated grasp poses.

However, our method cannot guarantee that all generated grasp poses are feasible. Certain conditions can render grasp poses invalid, primarily including (1) poor quality of the generated 3D bounding boxes and (2) generated grasp poses being too close to nearby objects. We believe that the performance can be further improved, for example, by annotating more 2D labels for the 3D detector or implementing additional filtering metrics. Figure 6 showcases some scenes with the grasp poses, including multiple objects, and occluded objects. 

\subsection{Comparing Grasping Poses}
In this experiment, we compare the grasp poses generated by our method with the grasp poses detected by 6dof GraspNet [18], GraspNet [20], and AnyGrasp [3] on the 1000 desktop scenes established in the section IV. We chose to compare on this dataset because our focus is on achieving robotic grasping without any grasp-specific training, which is a departure from the majority of existing grasp research that heavily relies on training. As a result, we directly apply existing methods to new scenes without retraining, simulating a scenario where grasp training is not required. This setting ensures a certain level of fairness in comparing our method to training-based approaches. However, comparing our method with existing approaches in new scenes presents a challenge in selecting appropriate evaluation metrics. When comparing training-based methods, different models are trained on the same dataset, allowing for the output of confidence for each grasp pose on the test set, thus enabling direct comparison with annotations. However, our method cannot follow this framework. Therefore, we introduce the stability metric (\textbf{M}), as defined in Eq.(6), as our comparative metric. Unlike the confidence generated by neural networks, this metric evaluates the stability of grasp poses based on structural properties such as object size and shape. As a result, it can be applied to measure grasp poses generated by any method.

\begin{figure*}[!t]
\centering
\includegraphics[width=0.8\linewidth]{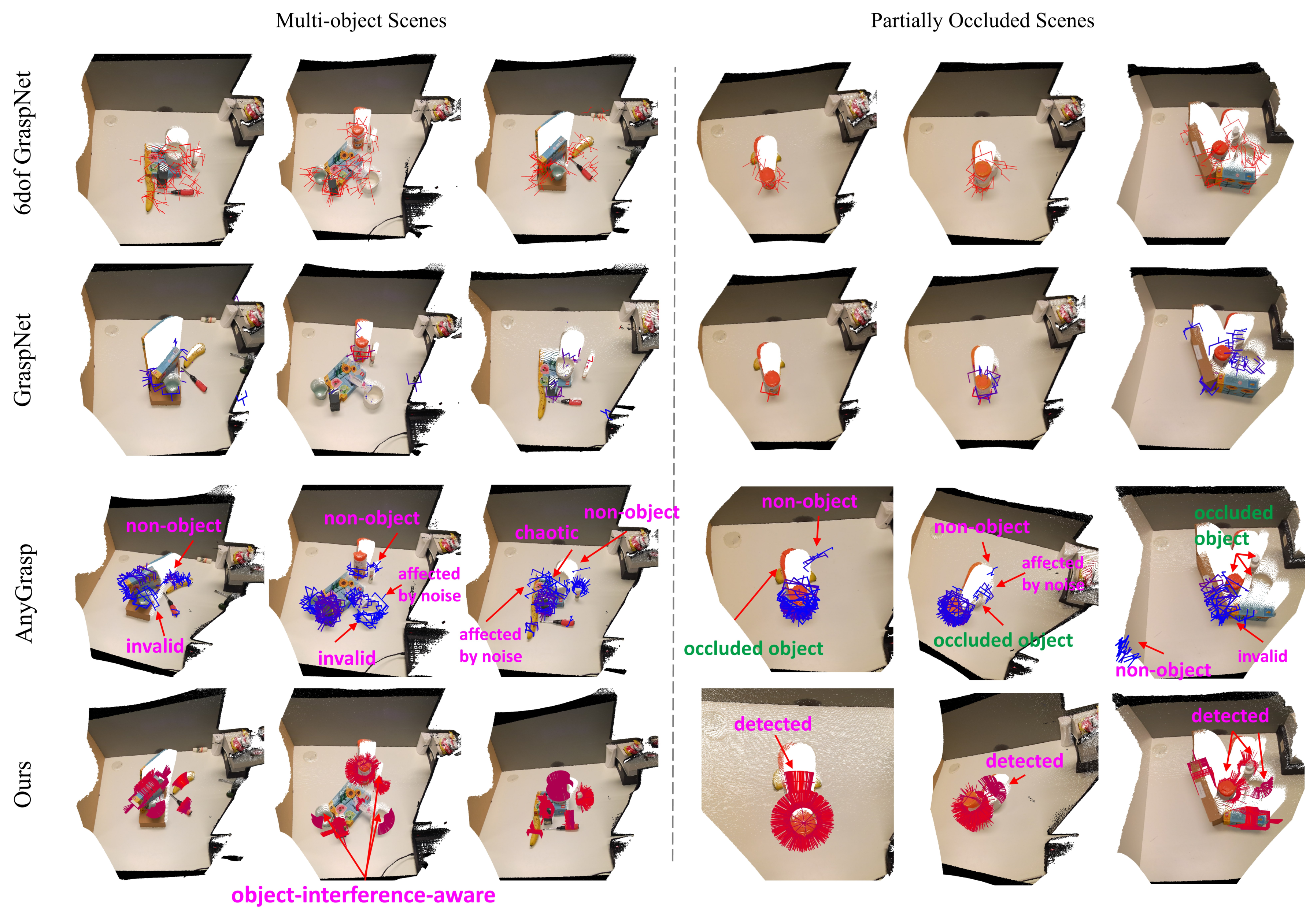}
\caption{The comparison of our method and other methods on multiple objects and partially occluded objects.}
\label{fig_2}
\end{figure*}

Our method directly generates grasp poses for each object in the scene and evaluates the stability of each grasp pose using the 3D bounding box of the object. However, other three methods cannot directly compute this value because the grasp poses generated by them are not classified, meaning it is not known which object each grasp pose corresponds to. Therefore, we manually perform statistics on the generated grasp poses and combined them with the 3D bounding boxes generated by our method to calculate the stability of the grasp poses. Specifically, we generate 100 grasp poses in each scene and then select up to three objects as statistical objects. For the selected objects, we rank the top five grasp poses and calculate the stability metric \textbf{M}. Then, we multiply \textbf{M} by a binary coefficient, denoted as $\beta$, where $\beta$ is set to 1 if the grasp pose is determined by the human to successfully grasp the object, and 0 otherwise. Finally, we calculate the mean value as the stability metric. Similarly, in our method, we select the same object in the same scene and utilize the same methodology to calculate the stability metric for the grasp poses.

\begin{figure}[!t]
\centering
\includegraphics[width=0.7\linewidth]{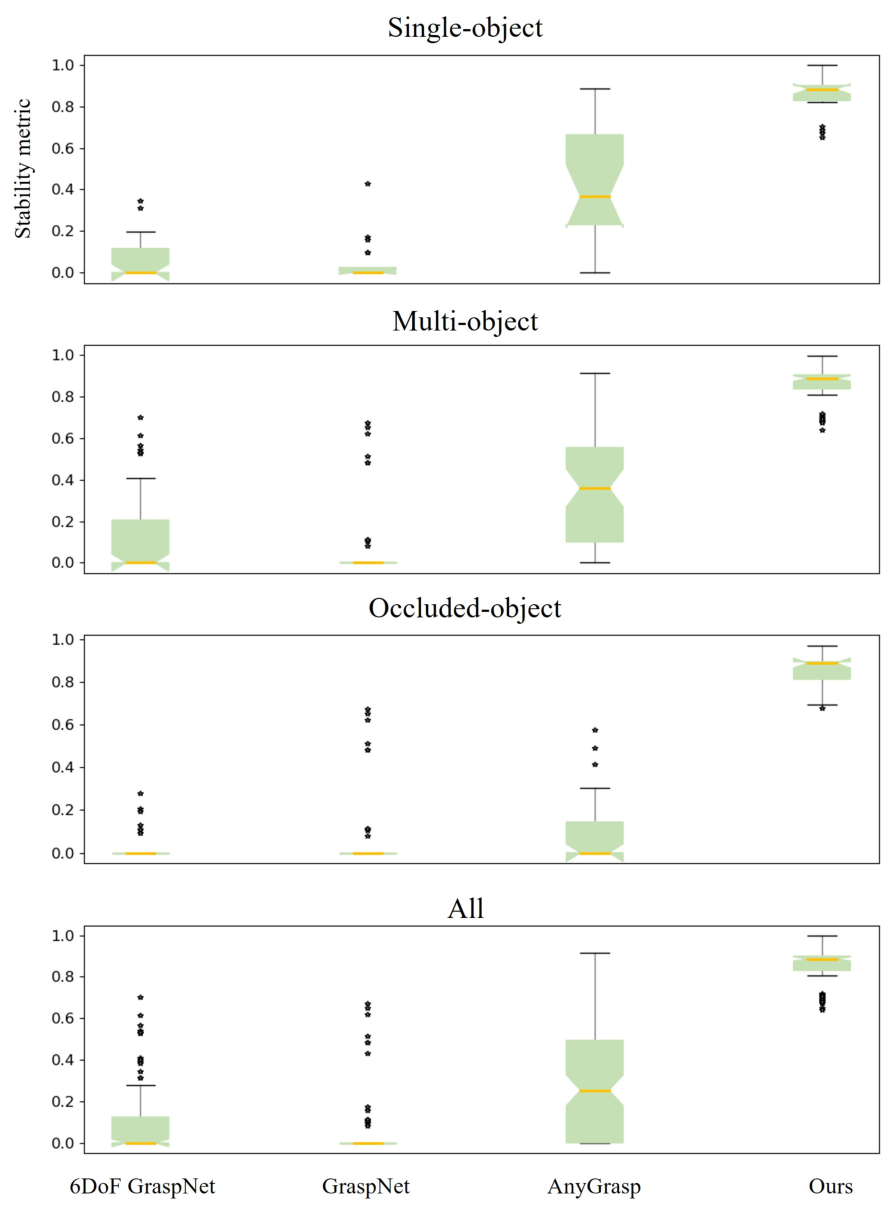}
\caption{The distributions of stability metric.}
\label{fig_2}
\end{figure}

Table I presents the grasp stability average values of our method, 6dof GraspNet, GraspNet, and AnyGrasp for 110 objects in 60 randomly selected scenes, including single objects, multi-objects, occluded objects. Please refer to Table II in the Supplementary Materials, where we have provided all the scenarios and detailed statistical data. The data demonstrate the grasp poses generated by our method exhibit significantly higher stability compared to other methods. 6dof GraspNet and GraspNet perform very poorly. We observe that AnyGrasp performs better in single-object scenes compared to multi-object scenes, possibly because it is more susceptible to interference from surrounding objects in multi-object scenarios. In occluded scenes, AnyGrasp struggles to generate effective grasp poses for occluded objects, indicating its inability to handle occlusion. Figure 7 demonstrates the notched box plots of our method and other methods. Our method exhibits significantly higher median values and a more concentrated data distribution, demonstrating the exceptional detection quality and robustness. 
Moreover, our method consistently demonstrates excellent performance across single-object, multi-object, and occluded scenarios. 

To provide a visual comparison, we visualize 6 scenes in Figure 6. To facilitate viewing, we reduce the number of grasp poses generated by our method. The grasp poses generated by our method demonstrate higher consistency and performance. More scenes can be accessed in the Supplementary Materials (Figure 7).

\subsection{Goal-oriented Grasping for A Scene}
In this section, we deploy GoalGrasp onto a real robot to accomplish the task of grasping user-specified targets. Specifically, the robot utilizes a depth sensor to capture scene data, and GoalGrasp is employed to generate grasp poses for each target. Upon receiving a grasping target instruction from the user, the robot executes the corresponding grasp for the specified target. In our experiments, we utilize an Ufactory xArm7 (7 DoF) and a xArm Gripper as the robot platform, along with a Microsoft Azure Kinect DK as the depth sensor. A computer equipped with an NVIDIA 3090 GPU is used to run the system.

\begin{table}[!t]
\caption{Stability metric in different scenarios. \textbf{Bold} is used to highlight the best results.}
\centering
\begin{tabular}{|c|c|c|c||c|}
\hline
Methods & Single & Multi-object & Occluded & Mean \\

\hline
6dof GraspNet & 0.069 & 0.126 & 0.037 & 0.077\\
GraspNet & 0.048 & 0.064 & 0 & 0.037\\
AnyGrasp & 0.412 & 0.362 & 0.086 & 0.287\\

\hline
\textbf{Ours} & \textbf{0.857} & \textbf{0.866} & \textbf{0.846} & \textbf{0.856}\\

\hline
\end{tabular}
\end{table}

\begin{figure}[!t]
\centering
\includegraphics[width=0.6\linewidth]{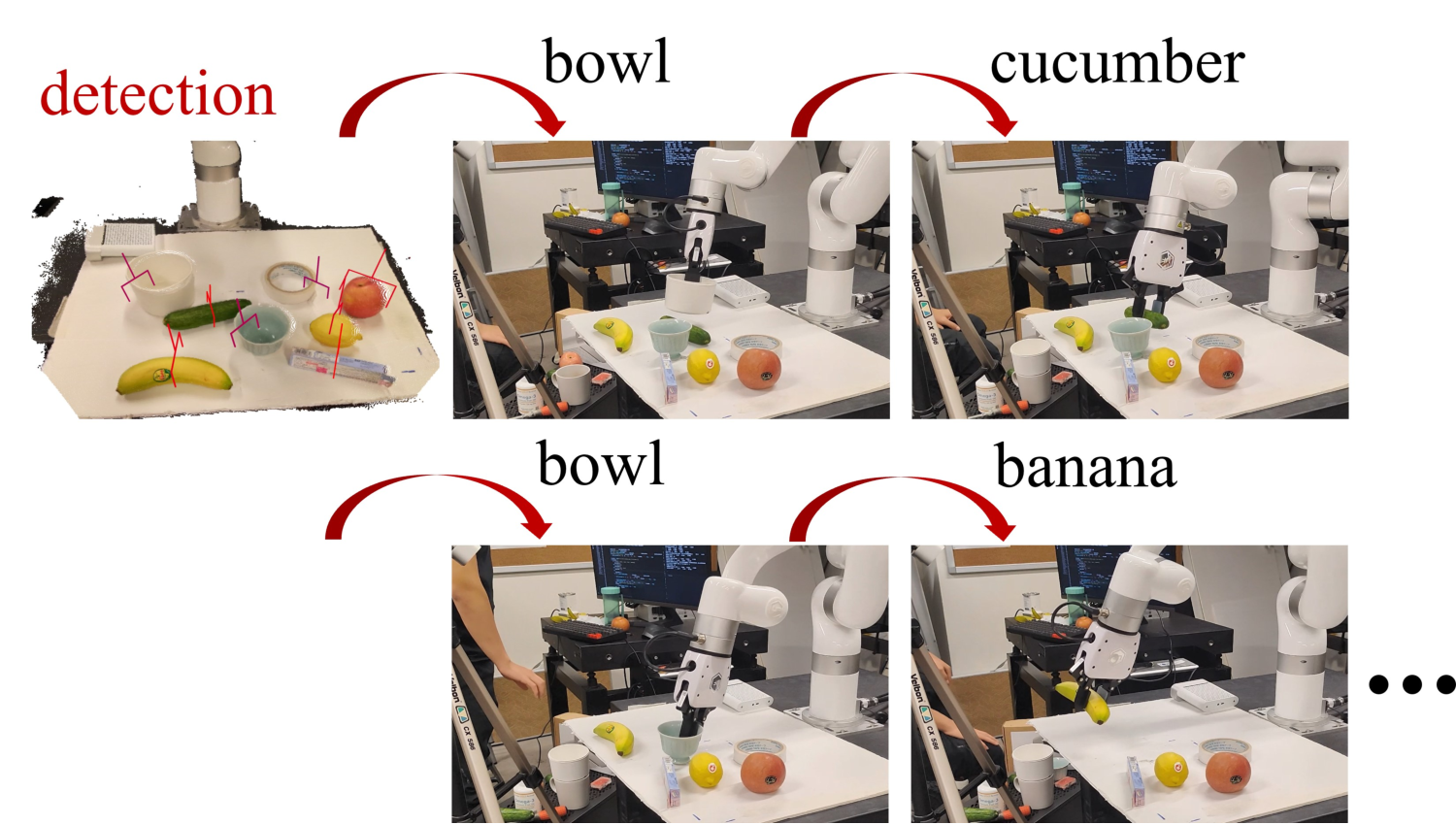}
\caption{User-specified grasping experiments, in which the object is determined by the user.}
\label{fig_2}
\end{figure}

We do not conduct experiments comparing other three methods with our method in this experimental setup because they cannot achieve target-oriented grasping, meaning it cannot complete the task of grasping user-specified objects. This target-oriented grasping is the focus of our work. Indeed, there are other studies on target-oriented robot grasping, but we also do not compare them with our method for the following reasons: (1) these studies mainly focus on top-down 3D grasping rather than 6D grasping, and (2) they do not demonstrate strong generalization capabilities to new scenes, making it difficult to guarantee their performance in novel scenarios.

GoalGrasp can operate in two modes: (1) First, it generates grasp poses for all objects in the scene and then waits for instructions. Once the instruction is received, the robot performs the grasping. After completing the grasp, it goes back to the waiting state for further instructions. This mode avoids the need to detect grasp poses for the target upon receiving an instruction, reducing the user's waiting time. However, the initial process of generating grasp poses for all objects in the scene can be time-consuming (about 1s). (2) The robot waits for the user's instruction. Upon receiving the instruction, it only detects the grasp poses for the target in the scene and performs the grasp. After completing the grasp, it goes back to the waiting for further instructions. This mode distributes the detection time across each grasping operation and can handle cases where object positions change. Table II presents the detection times for both modes. We select a scene with four objects for testing purposes. AnyGrasp performs four grasp detections, and after each detection, a human manually removes the corresponding target object. For GoalGrasp, we specify the grasping targets in the same order and conduct grasping trials, recording the time taken for detection in both two modes. Our method demonstrated a time for grasp detection approximately 50\% of that of AnyGrasp, indicating excellent real-time performance.

In our experiments, we conduct 500 grasping trials in both single-object and multi-object scenes (with a maximum of 8 objects), and the robot achieved a success rate of 94\%. Here, we define the success rate as the ratio of successful grasps to the total number of grasping attempts. Our method demonstrates a high success rate, which validates the effectiveness of our approach. Figure 8 illustrates the process of GoalGrasp generating a grasp pose for each object in the scene and the subsequent execution of the robot's grasping action based on user instructions. More experiments can be accessed in the Supplementary Materials (Figure 8). Additional experiment videos are available in \href{https://drive.google.com/file/d/1-5NEN8HcJd0lSAu6gPxk3ev0Yef9NWDO/view?usp=drive_link}{here}. This is an anonymous link.

\begin{table}[!t]
\caption{The Detection Time of Single Grasping Trial.Due to the limitations of AnyGrasp in achieving user-specified grasping, we only test the detection time but not execute the grasping. \label{tab:table1}}
\centering
\begin{tabular}{|c|c|c|c|c|c||c|}
\hline
 \multirow{2}{*}{\textbf{Methods}} & \multirow{2}{*}{\textbf{Modes}} & \multicolumn{4}{c|}{\textbf{Single detection (s)}} & \multirow{2}{*}{\textbf{Mean}}\\
\cline{3-6}
~&~&1&2&3&4&~\\

\hline

AnyGrasp & Mode \textbf{2} & 0.509 & 0.468 & 0.466 & 0.475 & 0.480\\
 
\hline
\multirow{2}{*}{GoalGrasp} & Mode \textbf{1} & 1.030 & \textbf{0} & \textbf{0} & \textbf{0} & 0.258\\
\cline{2-7}
 ~ &Mode \textbf{2} & \textbf{0.204} & 0.198 & 0.201 & 0.209 & \textbf{0.203} \\
\hline

\end{tabular}
\end{table}

\subsection{Grasping Partially Occluded Objects}
In this experiment, we validate that GoalGrasp can successfully grasp the target even when it is partially occluded (approximately 50\%). To the best of our knowledge, existing research has not achieved reliable grasping of target objects in partially occluded scenes. However, for home-service robots, this ability is essential. We employ the same setup as the previous experiment to evaluate this capability. Experimentally, we perform 100 grasping trials on occluded objects and achieve a grasp success rate of 92\% with the robot. Our method demonstrates high success rates in both single-object, multi-object, and partially occluded scenes, which indicates the robustness across different scenes. Figure 9 showcases one scene of the robot executing grasping tasks, providing perspectives from both the robot's perspective and a third-party viewpoint. More experiments can be accessed in the Supplementary Materials (Figure 9). The experiment videos are available in \href{https://drive.google.com/file/d/1-5nMJg0FeDHcEcYcW_YXFxTDy14aX_6R/view?usp=drive_link}{here}. This is an anonymous link.

\subsection{Success Rate on Real Robot}
We compare the grasp success rates of different methods on a real robot, where the success rate is defined as the proportion of successful grasps to the total number of grasp attempts. It is important to note that the experimental setups for our method and the compared methods are different. Our method performs grasps on user-specified objects from the scene, while the compared methods are tested in bin-picking scenarios, which means they are not tailored for grasping user-specified objects. To minimize the impact of different testing environments, we initially compared stabilization metrics as presented in Table I. Table III demonstrates the grasp success rates of our method and other methods on a real robot. It is essential to emphasize that this comparison is simplistic and aims to showcase that our method achieves relatively high grasp success rates. The experimental data for other methods is sourced from previously published papers [3,18].
We conducted these comparative experiments because existing methods are generally unable to grasp user-specified objects without relying on grasp training, making direct comparisons with our method challenging.

\begin{table}[!t]
\caption{Success rates of different methods on real robot.}
\centering
\begin{tabular}{|c|c||c|}
\hline
Methods & User-specified &  Success rates \\

\hline
6dof GraspNet & $\times$ & 88\% \\
AnyGrasp & $\times$ &  93.3\%\\

\hline
\textbf{Ours} & $\surd$ & 93.7\% \\

\hline
\end{tabular}
\end{table}

\section{Discussions and Future Work}

\subsection{New Objects}
Currently, our method follows the following steps for generating grasp poses for new objects: (1) categorizing objects based on their shapes, (2) designing corresponding grasp poses based on the object's 3D bounding box, category, and heuristics, and then generating the grasp poses. Indeed, heuristics guide the design of the grasp pose generation strategy. In our study, we have proposed several shape categories and classified 18 different objects into these categories. We utilize the corresponding heuristics and grasp pose generation algorithms to generate grasp poses, see Table III in the supplementary material. When encountering a new object, if it falls into one of the existing shape categories, we can directly apply the corresponding grasp pose algorithm. However, if the object cannot be categorized into existing shape categories, we need to derive human heuristics to design the corresponding grasp pose algorithm. As we consider a growing number of shape categories, our method will exhibit stronger scalability and flexibility. We acknowledge that currently we have not developed a systematic method for developing and extending existing grasping strategies across various object categories. This is primarily due to the diverse shapes of objects and the numerous ways humans grasp them, making it challenging to find a systematic solution.
Note that we do not have a clearly defined method for classifying objects. Currently, we rely on an experience-based classification approach. This is indeed a limitation of our current approach, and we are exploring new methods to address this issue. We are attempting to leverage large language models to classify objects and summarize grasping experiences, thereby fully automating the process.

Regarding heuristics, a similar challenge arises where it is difficult to find a generalized method to adapt to objects with different grasping patterns. 
The heuristic rules may be limitless. Currently, we employ straightforward rules; for example, for spherical objects, our rule (i.e., ST) is the entire sphere, which aligns with our intuition. How to manually derive these rules to ensure generalization is a profound issue and one we are actively researching. We view this as a study of the generalization capability of robotic grasping techniques from a logical perspective. In the future, we will explore the use of large language models to address this issue. By leveraging the strong logical capabilities of large language models, we summarize object categories and 3D bounding boxes into natural language inputs for the model, which then generates corresponding grasping rules. We apply these rules to grasp pose generation. Using the results from GoalGrasp, we can create instances for the large language model to guide its logical reasoning.

\subsection{Advantages and Limitations}
Our approach has two main advantages compared to existing methods. Firstly, it is capable of grasping user-specified targets, which requires our method to detect grasp poses with specific categories. This is challenging to achieve with current methods. Secondly, our approach addresses the issue of partial occlusion, as we have observed that existing methods struggle to detect effective grasp poses when the target object is partially occluded. One limitation of our method is that for new targets, manual annotation of 2D bounding boxes is required, along with the redesign of the grasp pose generation method if the new item does not belong to existing categories.

In the future, we plan to expand the collection of objects that can be grasped by incorporating more items. Additionally, we will integrate obstacle avoidance algorithms into the robotic system to further enhance the performance of grasping.

\begin{figure}[!t]
\centering
\includegraphics[width=0.8\linewidth]{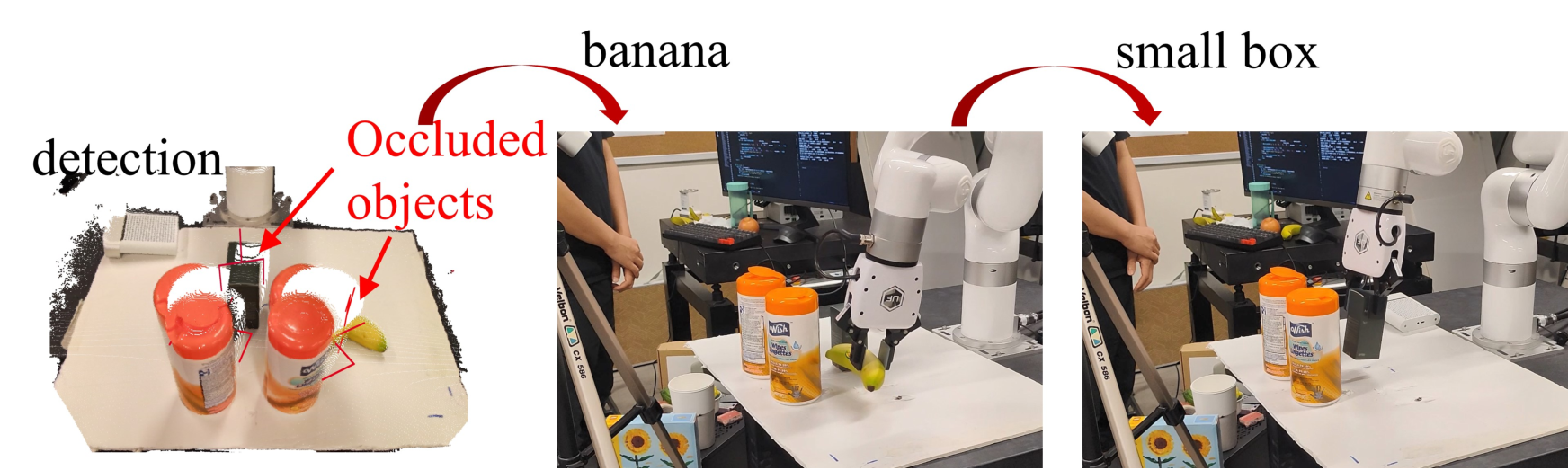}
\caption{User-specified grasping experiments in partially occluded scenarios.}
\label{fig_2}
\end{figure}

%\section*{ACKNOWLEDGMENT}
%Research supported by the Laboratory for Artificial Intelligence in Design (Project Code: RP1-3), Innovation and Technology Fund, Hong Kong Special Administrative Region

%%%%%%%%%%%%%%%%%%%%%%%%%%%%%%%%%%%%%%%%%%%%%%%%%%%%%%%%%%%%%%%%%%%%%%%%%%%%%%%%

\end{document}